\documentclass{article}

\usepackage[preprint]{neurips_2025}

\usepackage[utf8]{inputenc} 
\usepackage[T1]{fontenc}    
\usepackage{hyperref}       
\usepackage{url}            
\usepackage{booktabs}       
\usepackage{amsfonts}       
\usepackage{nicefrac}       
\usepackage{microtype}      
\usepackage[dvipsnames]{xcolor}         

\DeclareSymbolFont{bbold}{U}{bbold}{m}{n}
\DeclareSymbolFontAlphabet{\mathbbold}{bbold}

\usepackage{graphicx}
\usepackage{subfigure}
\usepackage{adjustbox}
\usepackage{comment}
\usepackage{tikz}
\usetikzlibrary{automata, positioning}

\usepackage{pdflscape}
\usepackage{amsmath}
\usepackage{amssymb}
\usepackage{mathtools}
\usepackage{amsthm}
\usepackage{multirow}
\usepackage{rotating}
\theoremstyle{plain}

\theoremstyle{definition}

\theoremstyle{remark}

\usepackage[textsize=tiny]{todonotes}

\usepackage{algorithmic}
\usepackage[nolist]{acronym}
\begin{acronym}[DRAM]
    \acro{AI}{Artificial Intelligence}
    \acrodefindefinite{AI}{an}{an}
    \acro{ASIC}{application-specific integrated circuit}
    \acrodefindefinite{ASIC}{an}{an}
    \acro{BD}{bounded delay}
    \acro{NN}{Neural Network}
    \acro{BNN}{Binarized Neural Network}
    \acro{CNN}{Convolutional Neural Network}
    \acro{FPTM}{Fuzzy-Pattern Tsetlin Machine}
    \acro{CTM}{Convolutional Tsetlin machine}
    \acro{CoTM}{Coalesced Tsetlin Machine}
    \acro{FSM}{Finite state machine}
    \acro{LA}{learning automaton}
    \acrodefplural{LA}{learning automata}
    \acrodefindefinite{LA}{an}{a}
    \acro{ML}{Machine Learning}
    \acrodefindefinite{ML}{an}{a}
    \acro{TA}{Tsetlin automaton}
    \acrodefplural{TA}{Tsetlin automata}
    \acro{TAT}{Tsetlin automaton team}
    \acro{TM}{Tsetlin Machine}
    \acro{GraphTM}{Graph Tsetlin Machine}
    \acro{GCN}{Graph Convolutional Neural Network}
    \acro{GNN}{Graph Neural Network}
\end{acronym}

\title{Fuzzy-Pattern Tsetlin Machine}

\author{%
  Artem Hnilov\thanks{Source code and additional resources are available at  \url{https://boobsd.github.io}} \\
  Independent Researcher \\
  \texttt{artem.hnilov@gmail.com} \\
}

\begin{document}

\maketitle

\begin{abstract}
The "\emph{all-or-nothing}" clause evaluation strategy is a core mechanism in the \ac{TM} family of algorithms. In this approach, each clause—a logical pattern composed of binary literals mapped to input data—is disqualified from voting if even a single literal fails. Due to this strict requirement, standard \ac{TM}s must employ thousands of clauses to achieve competitive accuracy. This paper introduces the \ac{FPTM}, a novel variant where clause evaluation is fuzzy rather than strict. If some literals in a clause fail, the remaining ones can still contribute to the overall vote with a proportionally reduced score. As a result, each clause effectively consists of sub-patterns that adapt individually to the input, enabling more flexible, efficient, and robust pattern matching. The proposed fuzzy mechanism significantly reduces the required number of clauses, memory footprint, and training time, while simultaneously improving accuracy. On the IMDb dataset, \ac{FPTM} achieves $90.15\%$ accuracy with only one clause per class, a $50\times$ reduction in clauses and memory over the Coalesced Tsetlin Machine. \ac{FPTM} trains up to $316\times$ faster ($45$ seconds vs. $4$ hours) and fits within $50$ KB, enabling online learning on microcontrollers. Inference throughput reaches $34.5$ million predictions/second ($51.4$ GB/s). On Fashion-MNIST, accuracy reaches $92.18\%$ ($2$ clauses), $93.19\%$ ($20$ clauses) and $94.68\%$ ($8000$ clauses), a $\sim400\times$ clause reduction compared to the Composite \ac{TM}'s $93.00\%$ ($8000$ clauses). On the Amazon Sales dataset with $20\%$ noise, \ac{FPTM} achieves $85.22\%$ accuracy, significantly outperforming the Graph Tsetlin Machine ($78.17\%$) and a Graph Convolutional Neural Network ($66.23\%$).
\end{abstract}

\section{Introduction}
\label{intro}

The \ac{TM} has emerged as a compelling paradigm in machine learning, offering distinct advantages in terms of interpretability and hardware efficiency~\cite{granmo2018tsetlin}. At its core, the \ac{TM} constructs logical, conjunctive clauses from binarized input features to learn complex patterns. A fundamental mechanism governing this process is the "\emph{all-or-nothing}" binary clause evaluation strategy. In this rigid approach, each clause, which represents a specific data pattern, contributes to the final classification decision only if every single one of its constituent literals aligns perfectly with the input data. Consequently, the failure of a single literal disqualifies the entire clause from voting on that input.

While effective, this strict evaluation requirement imposes a significant scalability challenge. To capture the inherent variability and nuance within complex datasets, standard \ac{TM}s must instantiate a massive number of clauses—often numbering in the thousands for each class. This proliferation of clauses leads to substantial computational overhead, demanding considerable memory resources to store the Tsetlin Automata states and significant time to converge during training. Variants such as the \ac{CoTM} have been developed to address some of these inefficiencies, yet they still rely on a large clause inventory to achieve competitive performance~\cite{glimsdal2021coalesced}.

This paper presents a novel and foundational enhancement to the \ac{TM} architecture: a fuzzy pattern-based Tsetlin Machine. The \ac{FPTM} replaces the conventional binary clause evaluation with a more
flexible, fuzzy evaluation mechanism. In the proposed model, a clause can still contribute to the overall classification vote even if a subset of its literals fails to match the input. The strength of its contribution is simply scaled proportionally to the number of literals that do match. Consequently, each clause in the \ac{FPTM} no longer represents a single monolithic pattern but rather a collection of \emph{adaptable sub-patterns} that are learned and evaluated individually. This approach facilitates a more robust, efficient, and granular form of pattern matching.

The implications of this fuzzy evaluation mechanism are profound. The \ac{FPTM} achieves a dramatic reduction in the number of clauses required, which translates directly into a smaller memory footprint and drastically accelerated training times, all while preserving high-level accuracy.

\section{Fuzzy-Pattern Clauses}

In standard \ac{TM}s, each clause contributes a binary vote—either $0$ or $1$—toward the final decision. In contrast, clauses in the \ac{FPTM} contribute votes ranging from $0$ up to a maximum value defined by a new hyperparameter, denoted as $LF$, which specifies the number of allowable literal failures within a clause before it is considered failed.

The core intuition behind this fuzzy evaluation is that a single \ac{FPTM} clause effectively represents a dense collection of \emph{sub-patterns}. In a standard \ac{TM}, a clause capturing the pattern "$A \land B \land C \land D$" fails if any literal is absent. In contrast, an \ac{FPTM} clause for the same pattern can still produce a strong vote for the sub-pattern "$A \land B \land D$" and a weaker vote for "$A \land C$". By scaling its contribution proportionally to the degree of match, each clause learns a primary pattern while inherently recognizing a combinatorial set of related, smaller patterns. This eliminates the need for the \ac{TM} to learn separate, near-identical clauses for every minor variation, leading to a more efficient and generalized pattern representation.

During evaluation, each clause initializes its vote count based on either the number of literals it contains or the value of $LF$, whichever is smaller. Specifically, if the number of literals is zero or exceeds $LF$, the vote count is set to $LF$; otherwise, it is initialized to the number of literals in the clause. This vote count is then decremented by one for each failed literal. If the number of failed literals exceeds the initialized vote count, the clause contributes zero votes. In other words, the vote count is clipped to zero if it becomes negative.

For example, if $LF = 50$, clause consist of $100$ literals and $15$ literals in a clause fail, the resulting vote is $50 - 15 = 35$. If $80$ literals fail, the clause contributes zero votes and is considered entirely failed. In a case where the clause contains only $20$ literals and $10$ of them fail, the effective vote limit is reduced to $20$ (instead of $50$), resulting in a final vote count of $10$.

This proportional voting is also the source of the \ac{FPTM}'s enhanced robustness to noise. An input feature corrupted by noise may cause one or two literals in a clause to fail, but it is unlikely to affect all of them. Where a standard \ac{TM} clause would be disqualified entirely, the fuzzy clause's vote is only marginally reduced, allowing the model to maintain predictive stability even with imperfect data.

In certain scenarios, it is necessary to explicitly determine whether a fuzzy clause has failed. A straightforward rule can be applied: if the number of votes resulting from clause evaluation is zero, the clause is considered to have \emph{failed entirely}. Conversely, if the number of votes is strictly greater than zero, the clause is regarded as having succeeded.

The special case where $LF = 1$ is algorithmically similar—but not identical—to the original \ac{TM} formulation (see below for further discussion). In this case, the clause evaluation logic is no longer fuzzy but instead strictly binary, as in all classical \ac{TM} variants.

\begin{figure}
    \centering
    \includegraphics[width=1\linewidth]{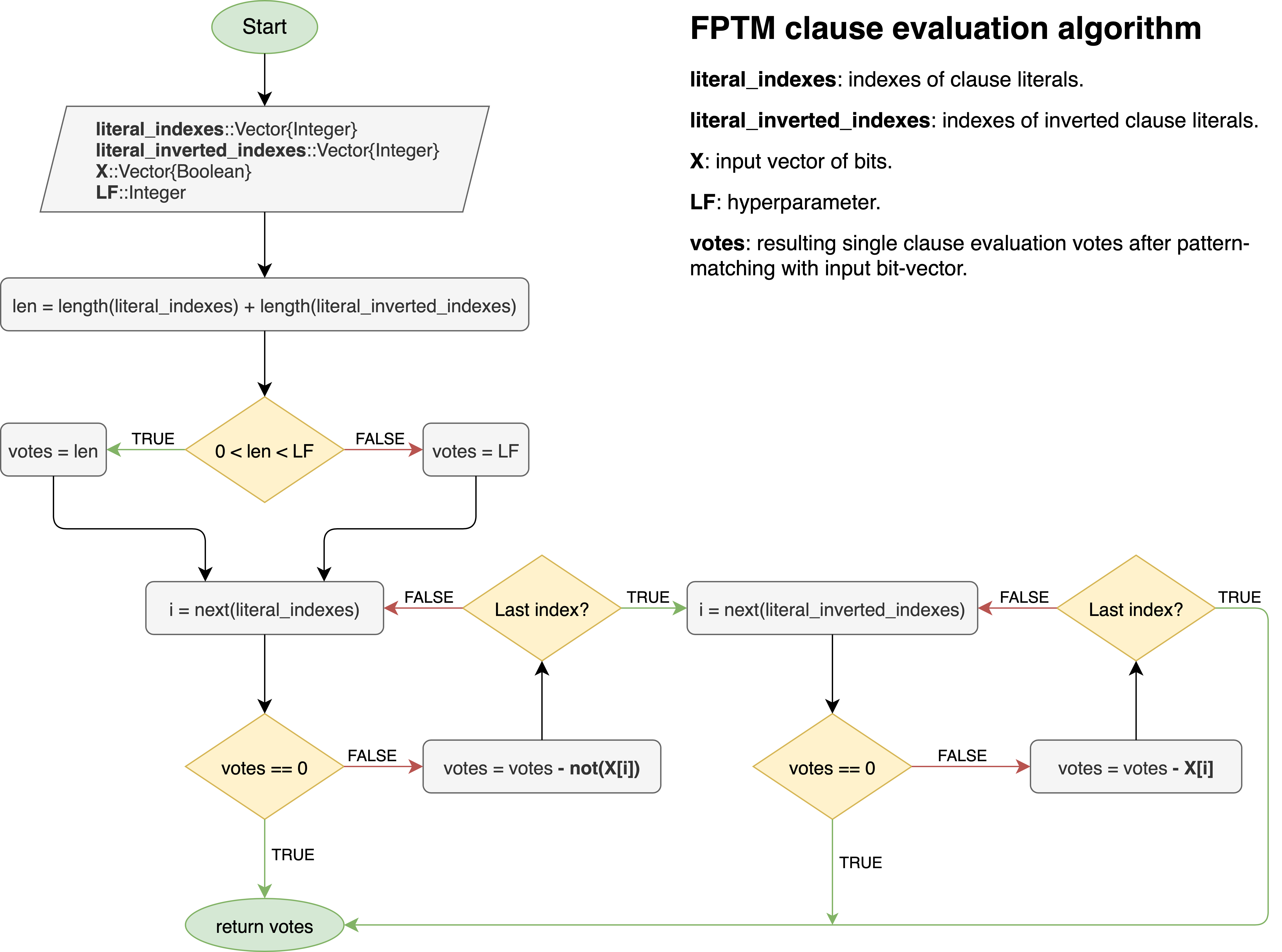}
    \caption{\ac{FPTM} single clause evaluation algorithm.}
    \label{fig:clause-eval-algo}
\end{figure}

The overall logic of single-clause evaluation in the \ac{FPTM} is illustrated in Figure \ref{fig:clause-eval-algo}.

Two other modifications from the standard \ac{TM} are introduced. First is a constraint on the maximum number of literals in a clause, controlled by the hyperparameter $L$. This is implemented in Type Ia feedback, wherein the Tsetlin Automata increment only if the number of literals in the clause is less than $L$, enforcing the clause size constraint~\cite{ijcai2023clause_size_limit}. Second, the Type Ia feedback mechanism is made deterministic, removing the probabilistic component of the feedback. The only remaining source of randomness corresponds to the hyperparameter $S$ in the Type Ib feedback. Due to this change, the value of $S$ is typically higher than in standard \ac{TM}s. The hyperparameter S is directly related to the Tsetlin Automaton's decrement probability in Type 1b feedback, defined by the inverse relationship:
$$P = \frac{1}{S}$$
Consequently, a larger value of $S$ reduces the frequency of state decrements in the Tsetlin Automata. The typically higher $S$ values employed in the \ac{FPTM} algorithm result in significantly fewer updates to the Tsetlin Automata state matrix, thereby accelerating the overall training process.

For multi-class classification, the \ac{FPTM} employs the classical \emph{one-vs-rest} strategy, wherein a separate binary classifier is trained for each class to distinguish that class from all others. However, this approach is not mandatory, and alternative methods can also be applied.

\subsection{Optimal hyperparameters}

Due to the different logic of clause evaluation, the optimal hyperparameters for the \ac{FPTM} differ from those of the standard \ac{TM}~\cite{olga2023optimal_hyperparameters}.

For multiclass \ac{FPTM}, the optimal value of the hyperparameter $T$ can be approximated as:

$$T \approx \sqrt{\frac{\text{CLAUSES}}{2} \times \text{LF}}$$

For binary \ac{FPTM}, the optimal $T$ is approximately:

$$T \approx \sqrt{\text{CLAUSES} \times \text{LF}}$$

where $\text{CLAUSES}$ denotes the number of clauses \emph{per class}, and $\text{LF}$ is a tunable hyperparameter.
In practice, the optimal value of $T$ may vary by a factor of two in either direction.

The optimal value of the $S$ hyperparameter for the \ac{FPTM} is approximately equal to the square of the optimal $S$ used in the Standard \ac{TM}:
$$S_{\text{FPTM}} \approx \left(S_{\text{TM}}\right)^2$$

This relationship appears to be logically linked to the elimination of randomness in the Type 1a feedback mechanism; however, further investigation is required to confirm this association.

The $LF$ hyperparameter is typically less than or equal to $L$, and is inversely proportional to the number of clauses ($\text{CLAUSES}$) per class. That is, the fewer the clauses in the model, the larger the value of $LF$ should be, and vice versa. In the case where $\text{LF} = 1$, the clauses are no longer fuzzy but instead become strict, following a classical "\emph{all-or-nothing}" binary evaluation strategy.

These approximations were determined empirically through extensive experimentation and provide a reliable starting point for hyperparameter tuning.

\section{Empirical Results}
\label{results}

The \ac{FPTM} was implemented from scratch in Julia (v1.11.6). All experiments were conducted on a consumer-grade desktop equipped with an AMD Ryzen 9 7950X3D CPU and 64 GB of DDR5 RAM (6000 MT/s), running Ubuntu 24.04.2 LTS. No GPU acceleration was utilized.

To evaluate the advantages of the \ac{FPTM}, three datasets were selected: IMDb~\cite{maas2011imdb}, Fashion-MNIST~\cite{xiao2017fmnist} and noisy Amazon Sales~\cite{amazonsales}. All source code is available online~\cite{FPTMcode}.

\subsection{IMDb Dataset}

The primary objective of this task was to test the \ac{FPTM} under extreme conditions, using only \textbf{one clause per class}, in contrast to traditional \ac{TM}, which typically employ hundreds or thousands of clauses. The baseline model was a Weighted \ac{CoTM} with a pool of 100 clauses ($\sim$50 per class). The parameters of the booleanization algorithm were identical across all models and are listed in Table \ref{imdb-params}.

\begin{table}
  \caption{Hyperparameters for IMDb dataset.}
  \label{imdb-params}
  \centering
  \begin{tabular}{rlll}
    \toprule
    Hyperparameters     & \ac{CoTM}     & Weighted \ac{CoTM} & Binary \ac{FPTM} \\
    \midrule
    Words number        & 40,000 & 40,000 & 40,000 \\
    Max n-gram           & 4     & 4     & 4     \\
    Features            & 12,800 & 12,800 & 12,800 \\
    \midrule
    Clauses per class   & $\sim$50 & $\sim$50 & 1 \\
    $T$                 & 80 & 80 & 18 \\
    $S$                 & 2 & 2 & 1,000 \\
    $L$                 &  &  & 64 \\
    $LF$                &  &  & 64 \\
    Weighted clauses    & no & yes & no \\
    Epochs              & 1,000 & 1,000 & 1,000 \\
    \bottomrule
  \end{tabular}
\end{table}

A total of 20 experiments were conducted using \ac{FPTM} and 4 experiments using \ac{CoTM} (due to significantly slower training), each spanning 1,000 epochs. The results are summarized in Table \ref{imdb-results}.

\begin{table}
  \caption{Empirical results for IMDb dataset.}
  \label{imdb-results}
  \centering
  \begin{tabular}{rlll}
    \toprule
    Result     & \ac{CoTM}     & Weighted \ac{CoTM} & Binary \ac{FPTM} \\
    \midrule
    Median peak test accuracy       & 88.47\% & \textbf{90.18\%} & 90.15\% \\
    Training time (1 thread)        & 27651 s & 14362 s & \textbf{399 s} \\
    Training time (32 threads)      & Not implemented & Not implemented & \textbf{45.5 s} \\
    Training speedup (1 thread)     & 0.52× & 1× (baseline) & \textbf{36×} \\
    Training speedup (32 threads)   & Not implemented & Not implemented & \textbf{316×} \\
    \bottomrule
  \end{tabular}
\end{table}

As observed, with comparable median peak test accuracy, \ac{FPTM} achieves a training speedup of \textbf{36}-fold on a single-threaded CPU and \textbf{316}-fold on a multi-threaded CPU compared to single-threaded Weighted \ac{CoTM}. Due to the lengthy runtime of \ac{CoTM}—approximately \textbf{four hours} per experiment—only four experiments were completed. In contrast, each \ac{FPTM} experiment required just \textbf{45 seconds}, enabling significantly more extensive experimentation.

For this task, the input data size is $12800$ bits, and each Tsetlin Automaton has $256$ states (one byte). In a 1-clause-per-class binary \ac{FPTM} model with negated features, the internal Tsetlin Automata state matrix occupies only:
$$\frac{12800 \times 2 \times 2}{1024} = 50\text{ KB}$$

In contrast, the \ac{CoTM} model occupies at least \textbf{50 times} more memory. This substantial reduction in memory usage opens up new possibilities for Tsetlin Machines—enabling the \emph{training} of relatively complex models on modern microcontrollers, which was previously infeasible.

Additionally, the resulting model was benchmarked using a batch inference technique inspired by the \emph{"REDRESS"} publication~\cite{maheshwari2023redress}. The model achieved \textbf{34.5 million} predictions per second in batch mode, corresponding to a throughput of \textbf{51.4 GB/s}, utilizing all available CPU cores.

\subsubsection{Optimal Model for the IMDb Dataset}

The IMDb dataset was also evaluated using four different configurations of the \ac{FPTM} model. This included one configuration with optimal hyperparameters and three with non-optimal hyperparameters. The purpose of this approach was to analyze the training behavior of the model under suboptimal settings. Table \ref{imdb-params-optimal} summarizes the parameters used in the booleanization algorithm and the corresponding hyperparameters. To clarify, unlike the previous experiment which employed the Binary \ac{FPTM}, this experiment utilized a multi-class classification model for the two-class IMDb dataset. This approach resulted in a slight improvement in accuracy compared to the Binary \ac{FPTM}, but it required twice as much memory.

\begin{table}
  \caption{Comparison of optimal and suboptimal hyperparameters of the \ac{FPTM} on the IMDb dataset.}
  \label{imdb-params-optimal}
  \centering
  \begin{tabular}{rllll}
    \toprule
    & \multicolumn{4}{c}{Multi-classification model}\\
    \cmidrule(lr){2-5}
    Hyperparameters
    & \ac{FPTM} Optimal & \ac{FPTM} (T = 320) & \ac{FPTM} (S = 200) & \ac{FPTM} (LF = 100) \\
    \midrule
    Words number        & 65,535 & 65,535 & 65,535 & 65,535 \\
    Max n-gram           & 4      & 4      & 4      & 4 \\
    Features            & 70,000 & 70,000 & 70,000 & 70,000 \\
    \midrule
    Clauses per class   & 200    & 200    & 200    & 200 \\
    $T$                 & 32     & \textbf{320}    & 32 & 32 \\
    $S$                 & 2,000  &  2,000 & \textbf{200} & 2,000 \\
    $L$                 & 100    & 100    & 100    & 100 \\
    $LF$                & 10     & 10     & 10     & \textbf{100} \\
    Epochs              & 200    & 200    & 200    & 200 \\
    \bottomrule
  \end{tabular}
\end{table}

\begin{figure}
    \centering
    \includegraphics[width=1\linewidth]{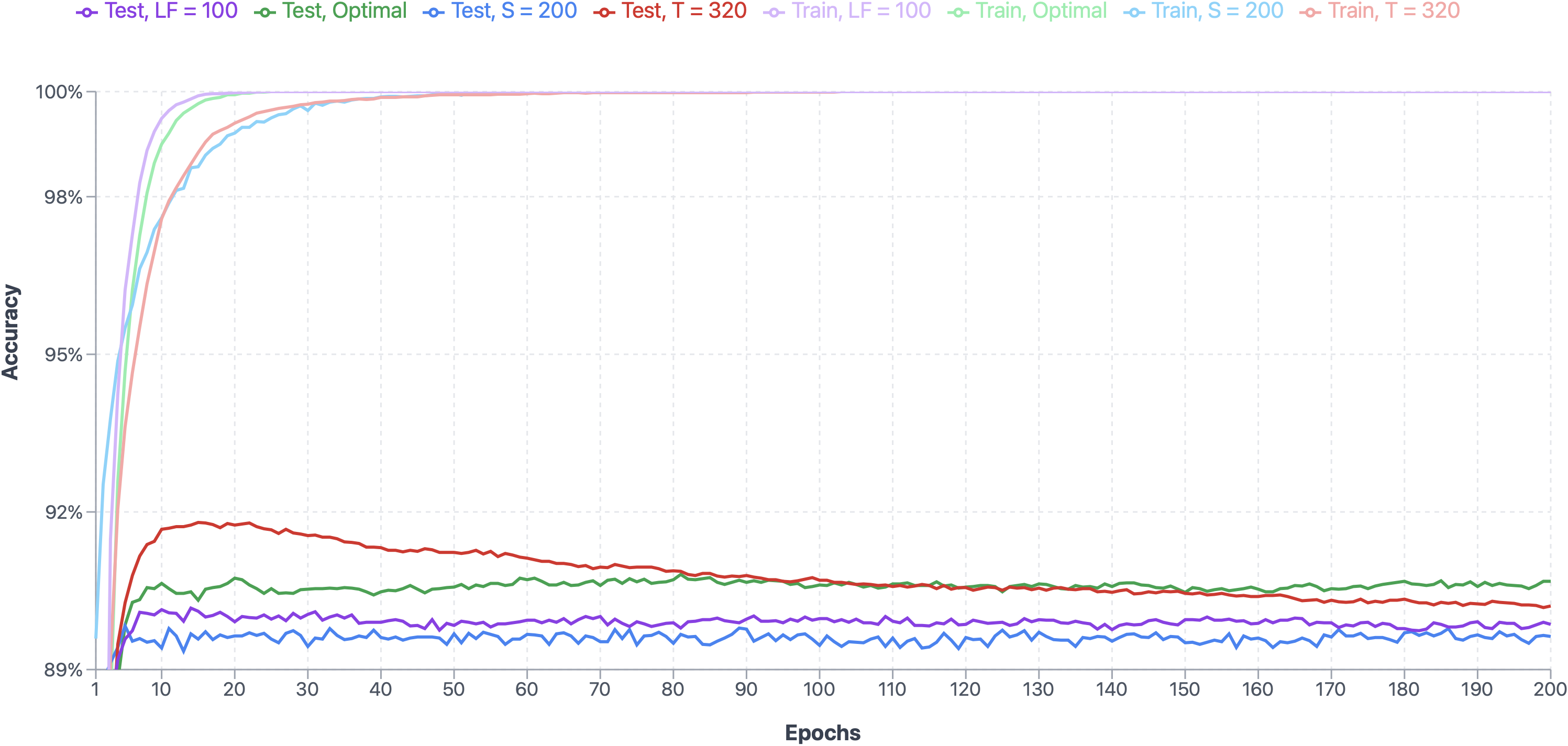}
    \caption{Training and testing accuracy comparison of the \ac{FPTM} model configured with optimal and suboptimal hyperparameters on the IMDb dataset.}
    \label{fig:imdb-optimal-accuracy}
\end{figure}

The training and testing accuracies are presented in Figure \ref{fig:imdb-optimal-accuracy}. As shown in the figure, the maximum test accuracy of $91.80\%$ is achieved at epoch 15 by a non-optimal model (with the hyperparameter $T$ set to 10 times the optimal value). However, after epoch 15, the testing accuracy begins to decline and continues to decrease in subsequent epochs, while the training accuracy continues to increase. This divergence is a strong indicator of overfitting. Based on this observation, it is possible to achieve a test accuracy of up to $91.70\%$ on the IMDb dataset by applying early stopping on the validation dataset after 15-20 epochs, a technique commonly used in neural network training.~\cite{prechelt1998early_stopping}.

In contrast, for other models—even with suboptimal hyperparameters $S$ and $LF$—no signs of overfitting are observed. The model with optimal hyperparameters achieves a test accuracy of $90.68\%$ after 200 epochs.

\begin{figure}
    \centering
    \includegraphics[width=1\linewidth]{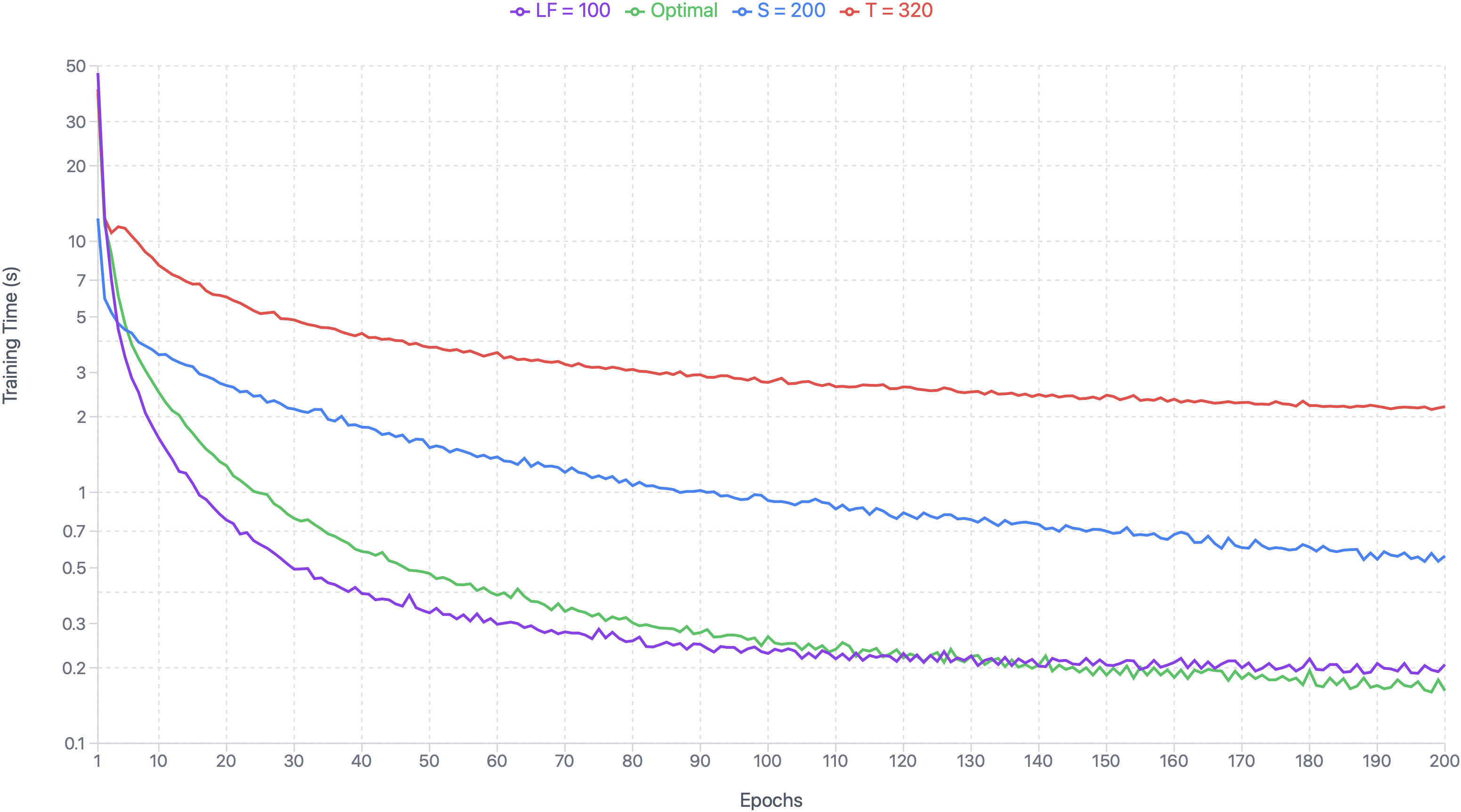}
    \caption{Training time comparison of the \ac{FPTM} model with optimal and suboptimal hyperparameters on the IMDb dataset (logarithmic scale).}
    \label{fig:imdb-optimal-training-time}
\end{figure}

The training times for all four models are presented in Figure \ref{fig:imdb-optimal-training-time}. As shown, the model with optimal hyperparameters achieves the fastest training over 200 epochs. In contrast, the slowest training time is observed in the model with suboptimal hyperparameter $T$, which reaches its maximum test accuracy after only 15 epochs.

\subsection{Fashion-MNIST Dataset}

The Fashion-MNIST dataset was evaluated using three different model configurations: multiclass \ac{FPTM} models with 2 and 20 clauses per class employing convolutional preprocessing, and a larger model comprising 8,000 clauses per class, augmented with additional data preprocessing techniques. Data augmentation involved standard techniques such as horizontal flips, small rotations and zoom transformations. It is important to note that the convolutional preprocessing is applied during the booleanization stage and is not an inherent component of the \ac{FPTM} itself~\cite{amini2016accelerated}. Several fixed convolutional kernels ($3\times3$, $5\times5$, and $7\times7$) were used for edge detection. The resulting feature maps were thresholded based on the pixel value histogram and concatenated into a single boolean bit vector. Each configuration was trained over 1,000 epochs across multiple experimental runs. The results are summarized in Table \ref{fashion-mnist-results}.

\begin{table}
  \caption{Hyperparameters and test accuracies after 1,000 training epochs for \ac{FPTM} models of varying sizes on the Fashion-MNIST dataset.}
  \label{fashion-mnist-results}
  \centering
  \begin{tabular}{rlll}
    \toprule
    & \multicolumn{3}{c}{Model}\\
    \cmidrule(lr){2-4}
    & \ac{FPTM} tiny & \ac{FPTM} small & \ac{FPTM} big \\
    \midrule
    Clauses per class & 2 & 20 & 8,000  \\
    $T$             & 80 & 100 & 700 \\
    $S$             & 1,000 & 700 & 700 \\
    $L$             & 1,200 & 200 & 30 \\
    $LF$            & 1,200 & 200 & 30 \\
    Augmented data  & no & no & yes \\
    Test accuracy   & 92.18\% & 93.19\% & 94.68\% \\
    \bottomrule
  \end{tabular}
\end{table}

The highest previously reported accuracy for the Composite Convolutional \ac{TM} on the Fashion-MNIST dataset is \textbf{93.00\%}, achieved using \textbf{8,000} clauses per class with a similar compositional approach~\cite{granmo2023tmcomposites}. In contrast, \ac{FPTM} attained the same level of accuracy using only \textbf{20} clauses per class—representing a \textbf{400×} reduction in clause count. This significant reduction also leads to substantially faster training, with the 1,000-epoch training process completing in just 10 minutes on desktop CPU.

A test accuracy of \textbf{94.68\%} establishes a new \emph{state-of-the-art} among all \ac{TM} variants, demonstrating its capability to outperform complex neural network architectures, including \emph{Inception-v3} with specialized convolutional filters, as reported in~\cite{gavrikov2022inception}.

\subsection{Noisy Amazon Sales Dataset}

Finally, the Amazon Sales dataset, evaluated under six different noise ratios, was used to assess the performance of the proposed \ac{FPTM} model for a recommendation system task. In this study, the modeling framework is designed to capture the interdependencies among three key entities in the dataset: customers, products, and their associated categories. The \ac{FPTM} model is trained on supervised data consisting of user-provided ratings or preference scores, which serve as labels indicating the degree of user satisfaction or interest associated with specific customer–product–category triplets.

The results were compared with those of the standard \ac{TM}, the \ac{GCN}, and the more recent \ac{GraphTM}~\cite{granmo2025graphtm}. Hyperparameters for noisy Amazon Sales dataset can be found in Table \ref{amazon-sales-params}.

\begin{table}
  \caption{FPTM model hyperparameters for the noisy Amazon Sales dataset.}
  \label{amazon-sales-params}
  \centering
  \begin{tabular}{rlll}
    \toprule
    Hyperparameters     & Value \\
    \midrule
    Clauses per class   & 2,000 \\
    $T$                 & 100 \\
    $S$                 & 1,000 \\
    $L$                 & 50 \\
    $LF$                & 10 \\
    \bottomrule
  \end{tabular}
\end{table}

The \ac{FPTM} model was trained for 1,000 epochs in multiple experiments conducted for each noise level of the dataset. \ac{FPTM} consistently outperformed all competing approaches across all noise ratios. A summary of the results is provided in Table \ref{amazon-sales-results}.

\begin{table}
  \caption{Test accuracies of the Standard \ac{TM}, \ac{GCN}, \ac{GraphTM}, and \ac{FPTM} models on the noisy Amazon Sales dataset under varying noise ratios.}
  \label{amazon-sales-results}
  \centering
  \begin{tabular}{rllllll}
    \toprule
    & \multicolumn{6}{c}{Noise ratio} \\
    \cmidrule(r){2-7}
    Model               & 0.005 & 0.01 & 0.02 & 0.05 & 0.1 & 0.2 \\
    \midrule
    Standard \ac{TM}    & 76.64 & 74.86 & 72.40 & 64.00 & 49.35 & 20.12 \\
    \ac{GCN}            & 99.07 & 91.56 & 90.96 & 84.84 & 70.87 & 66.23 \\
    \ac{GraphTM}        & 98.63 & 98.42 & 97.84 & 94.67 & 89.86 & 78.17 \\
    \ac{FPTM}           & \textbf{99.64} & \textbf{99.07} & \textbf{98.72} & \textbf{96.01} & \textbf{91.56} & \textbf{85.22} \\
    \bottomrule
  \end{tabular}
\end{table}

These strong results—particularly at the $20\%$ noise level—demonstrate that \ac{FPTM} is capable of effectively handling noisy data. This robustness can be attributed to the flexible fuzzy clause evaluation mechanism employed by \ac{FPTM}, in which each clause consists of a combinatorial collection of \emph{adaptable sub-patterns} that are learned and evaluated individually. Further investigation of this behavior under noisy conditions across different datasets is warranted. However, this remains a topic for future research.

\section{Conclusions}
\label{conclusions}

This paper introduces the \ac{FPTM}, a novel architecture that marks a paradigm shift in the \ac{TM} family of algorithms. By replacing the traditional, rigid "\emph{all-or-nothing}" binary clause evaluation with a flexible fuzzy mechanism, the \ac{FPTM} learns robust and compact representations that significantly enhance performance across multiple domains.

The empirical evaluation demonstrates the profound impact of this new approach. On the IMDb dataset, the \ac{FPTM} achieved a median peak test accuracy of $90.15\%$ using only a single clause per class, representing a $50$-fold reduction in both clause count and memory usage compared to the Weighted \ac{CoTM}. This remarkable efficiency, which reduces the Tsetlin Automata state matrix to just $50$ KB, makes potential online training on microcontrollers a practical reality for the first time. The training process is accelerated by up to $316\times$, cutting the time for $1000$ epochs from four hours to a mere $45$ seconds while maintaining comparable accuracy.

On the Fashion-MNIST dataset, the \ac{FPTM} achieved a new \emph{state-of-the-art} accuracy of $94.68\%$ among all \ac{TM} variants. Additionally, a smaller \ac{FPTM} model attained $93.19\%$ accuracy while reducing the number of clauses by a factor of $400$ compared to the previous Composite \ac{TM} approach with comparable performance. Furthermore, its performance surpasses that of complex neural network architectures like \emph{Inception-v3} on this benchmark. The \ac{FPTM} also demonstrated superior robustness on the noisy Amazon Sales dataset, outperforming the \ac{GCN} and latest \ac{GraphTM} models across all noise levels, achieving an impressive $85.22\%$ accuracy even with $20\%$ noise.

While the empirical results are compelling, we acknowledge that the introduction of the $LF$ hyperparameter adds a new dimension to model tuning, and the relationships between hyperparameters are currently guided by empirical observation. Future work should focus on developing a more formal theoretical framework to analyze the convergence properties of fuzzy clauses. Investigating adaptive methods to automatically learn the optimal $LF$ value during training would also be a valuable research direction.

In conclusion, the \ac{FPTM} addresses the most significant scalability and efficiency challenges of traditional \ac{TM}s. By enabling more granular and robust pattern matching, the \ac{FPTM} stands as a faster, smaller, and more robust algorithm, unlocking new possibilities for interpretable machine learning in both high-performance and resource-constrained environments.

\bibliography{references}
\bibliographystyle{unsrt}


\end{document}